\documentclass[11pt]{article}

\PassOptionsToPackage{table}{xcolor}
\usepackage[preprint]{acl}
\usepackage{times}
\usepackage{latexsym}
\usepackage[T1]{fontenc}
\usepackage[utf8]{inputenc}
\usepackage{microtype}
\usepackage{inconsolata}
\usepackage{graphicx}
\usepackage{booktabs}
\usepackage{amsfonts}
\usepackage{array}
\usepackage{tabularx}
\usepackage{caption}
\usepackage[most]{tcolorbox}
\usepackage{tikz}
\usepackage{xurl}
\usetikzlibrary{arrows.meta,positioning,calc}

\newcommand{\evalcardd}{\texttt{EvalCard}}
\newcommand{\evalcards}{\texttt{EvalCards}}
\newcommand{\para}[1]{\paragraph{#1}}

\definecolor{pastelblue}{RGB}{204,229,255}
\definecolor{pastelpink}{RGB}{255,204,229}
\definecolor{lightgray}{gray}{0.95}
\definecolor{azure(colorwheel)}{rgb}{0.0, 0.5, 1.0}

\renewcommand{\arraystretch}{1.1}

\newtcolorbox{evalcardv2}[1][]{
  enhanced,
  breakable,
  colback=white,
  colframe=azure(colorwheel),
  title=\textbf{EvalCard: #1},
  fonttitle=\bfseries,
  coltitle=white,
  arc=8pt,
  width=14.75cm,
  boxrule=1pt,
  before skip=2pt, after skip=2pt,
  attach boxed title to top left={
    yshift=-3mm, xshift=5mm
  },
  boxed title style={
    size=small,
    colback=azure(colorwheel),
    colframe=azure(colorwheel),
    sharp corners=south
  },
}

\title{EvalCards: A Framework for Standardized Evaluation Reporting}

\author{
Ruchira Dhar$^{1}$\thanks{Primary author. All correspondence to \texttt{rudh@di.ku.dk}.} \quad \\
Danae S\'anchez Villegas$^{1}$ \quad
Antonia Karamolegkou$^{1}$ \\
Alice Schiavone$^{1}$ \quad
Yifei Yuan$^{2}$ \quad
Xinyi Chen$^{3}$ \quad
Jiaang Li$^{1}$ \\
Stella Frank$^{1}$ \quad
Laura De Grazia$^{4}$ \quad
Monorama Swain$^{5}$ \\
Stephanie Brandl$^{1}$ \quad
Daniel Hershcovich$^{1}$ \\
Anders S{\o}gaard$^{1}$ \quad
Desmond Elliott$^{1}$ \\[0.2cm]
\normalfont $^{1}$ University of Copenhagen \quad $^{2}$ ETH Zurich \\
\normalfont $^{3}$ University of Amsterdam \quad $^{4}$ University of Barcelona \\
\normalfont $^{5}$ Johannes Kepler University Linz
}

\begin{document}

\maketitle

\begin{abstract}
  Evaluation reports guide the use of general-purpose AI (GPAI) models, yet inconsistent reporting often leaves essential questions unanswered. Through a review of prior work, we identify three central issues in evaluation reporting practices today, \textit{reproducibility}, \textit{accessibility}, and \textit{accountability}, and examine how existing reporting artifacts fall short in addressing them. We introduce the EvalCards framework, outlining its key dimensions and presenting a preliminary format for concise evaluation summaries. We conclude by discussing how the framework can guide the future development of evaluation reporting standards.

\end{abstract}

\section{Introduction}

\begin{figure}[t]
    \centering
    \includegraphics[width=\columnwidth]{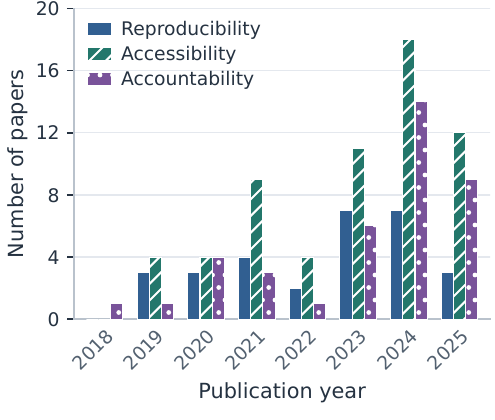}
    \caption{Tagged corpus with three-way reporting issue classification based on \autoref{tab:issuedefinitions}, grouped by publication year. Each paper supports multiple tags.}
    \label{fig:synthesistags}
\end{figure}

General-purpose AI (GPAI) models are increasingly used to support decisions with real-world consequences and evaluations can provide key evidence for responsible deployment, helping stakeholders assess whether a model is suitable for a particular use \cite{laskar-etal-2024-systematic}. Yet, prior work has shown how essential details about what was tested and under which conditions are often missing or scattered across sources, making reported evaluation results difficult to verify and interpret \cite{burnell2023rethink, biderman2024lessons}. Alongside these critiques, the research community has also proposed approaches to standardize documentation about evaluation:  Model Cards \cite{mitchell2019model} and FactSheets \cite{arnold2019factsheets} structure disclosures about models and AI services, while Datasheets \cite{gebru2021datasheets} and BenchmarkCards \cite{sokol2024benchmarkcards} document datasets and benchmarks. These contributions establish valuable reporting practices, but leave open how the findings on reporting problems and proposed remedies can directly inform a coherent framework for GPAI evaluation reporting. Our work aims to address this gap.

\definecolor{reprocolor}{HTML}{315F91}
\definecolor{accesscolor}{HTML}{23776B}
\definecolor{accountcolor}{HTML}{79539A}

\providecommand{\evalcite}[2]{}
\renewcommand{\evalcite}[2]{%
  \begingroup
  \hypersetup{citecolor=#1}%
  \tcbox[
    on line,
    colback=#1!8!white,
    colframe=#1!20!white,
    coltext=#1,
    boxrule=0.25pt,
    arc=2pt,
    boxsep=0pt,
    left=3pt, right=3pt,
    top=1.5pt, bottom=1.5pt
  ]{\scriptsize\strut\citet{#2}}%
  \endgroup
}

\begin{table*}[t]
\centering
\small
\setlength{\tabcolsep}{5pt}
\renewcommand{\arraystretch}{1.15}

\begin{tabularx}{\textwidth}{
  @{}
  >{\raggedright\arraybackslash}p{0.17\textwidth}
  >{\raggedright\arraybackslash}X
  >{\raggedright\arraybackslash}p{0.39\textwidth}
  @{}
}
\toprule
\textbf{Issue} &
\textbf{Definition} &
\textbf{Representative works} \\
\midrule

\textbf{Reproducibility} &
The gaps in methodological reporting that limit stakeholders'
ability to reproduce or independently verify evaluation results. &
\evalcite{reprocolor}{belz-etal-2021-systematic},
\evalcite{reprocolor}{palen-michel-etal-2021-seqscore},
\evalcite{reprocolor}{storks-etal-2023-nlp},
\evalcite{reprocolor}{belz-etal-2023-non},
\evalcite{reprocolor}{laskar-etal-2024-systematic},
\evalcite{reprocolor}{biderman2024lessons}
\\[6pt]

\textbf{Accessibility} &
The barriers to finding and interpreting evaluation results
that limit stakeholders' ability to make informed comparisons
and decisions. &
\evalcite{accesscolor}{ethayarajh-jurafsky-2020-utility},
\evalcite{accesscolor}{crisan2022interactive},
\evalcite{accesscolor}{liao2023designerly},
\evalcite{accesscolor}{burnell2023rethink},
\evalcite{accesscolor}{sokol2024benchmarkcards},
\evalcite{accesscolor}{zhao2025workflows}
\\[6pt]

\textbf{Accountability} &
The gaps in evaluation evidence and mechanisms for challenge and response
that limit stakeholders' ability to assess model risks and exercise oversight. &
\evalcite{accountcolor}{raji2020closing},
\evalcite{accountcolor}{wieringa2020account},
\evalcite{accountcolor}{cattell2024coordinated},
\evalcite{accountcolor}{weidinger2024holistic},
\evalcite{accountcolor}{mcgregor2025err},
\evalcite{accountcolor}{longpre2025position}
\\

\bottomrule
\end{tabularx}

\caption{Three umbrella issues in evaluation reporting identified through critical synthesis.}
\label{tab:issuedefinitions}
\end{table*}

Through a review and critical synthesis of relevant work in the area, we identify three dimensions of issues in evaluation reporting: \textit{reproducibility}, \textit{accessibility}, and \textit{accountability}, and examine how established reporting artifacts leave gaps in suitability for reporting GPAI model evaluations. We bring these findings together in \evalcards{}, a guiding framework based on the three dimensions and also present a preliminary artifact illustrated through three GPAI model case studies \footnote{The survey materials and code for generating the artifact will be publicly released upon publication.}. The framework provides a foundation for translating the research community's accumulated insights into evaluation reporting standards and artifacts for GPAI models.

\section{Issues in Evaluation Reporting}
\label{sec:currentstate}

To identify common issues in evaluation reporting of artificial intelligence (AI) models, including GPAI, we conduct a literature review and critical synthesis following the approach from similar works \cite{shelby2023sociotechnical, eriksson2025can}, supplementing the snowball sampling with keyword searches for corpus creation and structuring our synthesis around reporting shortcomings and their consequences for stakeholders.

\para{Literature Review}\  We use seminal works in evaluation reporting of models as seeds \cite{belz-etal-2021-systematic, ott2022mapping, burnell2023rethink, belz-etal-2023-non,laskar2024systematic, biderman2024lessons, reuel2025evaluates}, followed by backward and forward snowball sampling.  We also supplement this with systematic keyword searches (\textit{evaluation report}, \textit{model report}, \textit{evaluation disclosure}) in ACL Anthology, DBLP, and Google Scholar. We include works published through 2025 that: i) identify evaluation reporting problems or ii) examine consequences of such reporting problems or iii) propose new evaluation reporting practices. We imposed no application-domain restriction, provided that a paper met at least one inclusion criterion. We excluded papers concerned solely with developing evaluation methods, benchmarks, metrics, or human annotation protocols without substantive discussion of evaluation reporting. The retained corpus comprises 68 papers published between 2018 and 2025. 

\para{Critical Synthesis} For each paper, we annotate (RQ1) the evaluation-reporting shortcomings identified and (RQ2) their consequences for stakeholder activities. Following thematic synthesis \cite{thomas2008methods} and related review approaches \cite{eriksson2025can,shelby2023sociotechnical}, all authors iteratively code and group these annotations, refining category definitions through discussion. Paper-level annotations and category assignments are prepared by one author and reviewed by two others, with disagreements resolved through discussion. This process yields three dimensions: reproducibility, accessibility, and accountability (\autoref{tab:issuedefinitions}). We assign all applicable categories to each paper, allowing overlap across dimensions. \autoref{fig:synthesistags} summarizes their distribution by publication year. Next, we discuss the issues identified under each dimension in more details.

\subsection{Reproducibility Issues}
\label{sec:reproducibility}

This issue concerns how missing methodological details prevent stakeholders from independently verifying evaluation reports.

\para{Capability and benchmark specification.}\ Reports should identify the capabilities and modalities evaluated and link them to the relevant benchmarks \cite{carro2025conceptual, laskar-etal-2024-systematic}. Without these specifications, stakeholders may be unable to reconstruct the scope of claimed capabilities \cite{arnold2019factsheets, hernandez2020minimar, biderman2024lessons}.

\begin{figure*}[t]
  \centering
  \includegraphics[width=\textwidth, trim={3.7cm 4.0cm 3.7cm 6cm}, clip]{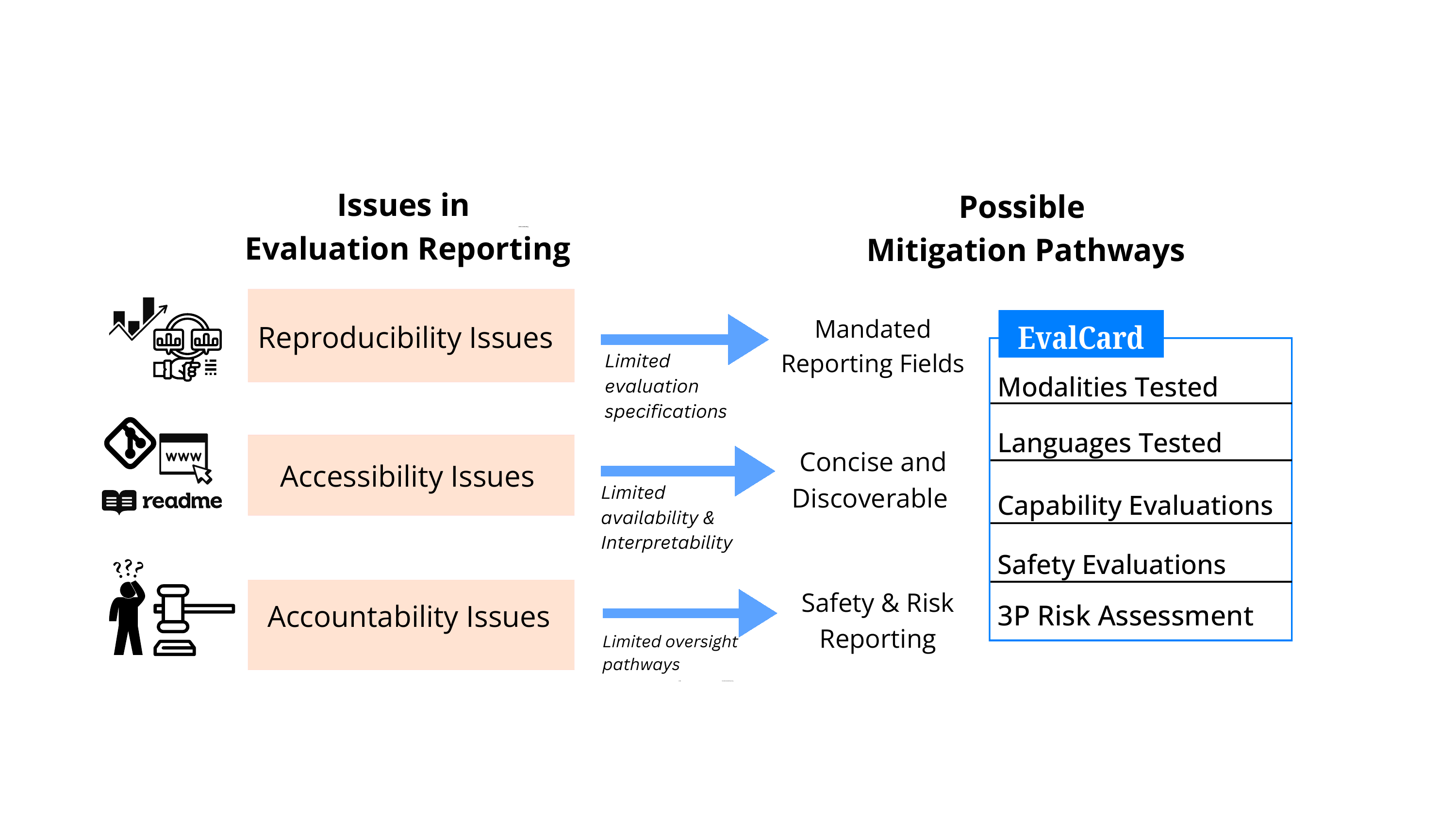}
  \caption{Issues in evaluation reporting and possible mitigation pathways for each, offering design directions for future reporting artifacts.}
  \label{fig:evalfig}
\end{figure*}

\para{Prompting specification.}\ Evaluation scores for GPAI models can vary significantly with prompt configurations provided to a model \cite{biderman2024lessons, laskar-etal-2024-systematic}. Reports should therefore document prompting methods needed to reproduce the evaluation \cite{park2024reporting, biderman2024lessons}.

\para{Metric specification.}\ Reports should identify the metrics used and specify how model outputs are processed, scored, and aggregated \cite{palen-michel-etal-2021-seqscore, laskar-etal-2024-systematic, biderman2024lessons}. For human evaluations, reproductions also require clear assessment criteria, evaluator instructions, and procedural details \cite{belz-etal-2023-missing, belz-etal-2023-non, liu-etal-2023-responsible}.

\smallskip
\noindent
$\rightarrow$ \textit{What we need:} Evaluation reports should specify the evaluated capabilities, link them to specific benchmarks, and associate each metric score with its prompting configuration.

\subsection{Accessibility Issues}

This issue centers around stakeholders' ability to find and interpret evaluation reports for informed decision-making.

\para{Limited availability.}\ Evaluation results and supporting details are not always publicly available \cite{haibe2020transparency, burnell2023rethink, belz-etal-2023-missing}. When disclosed, they may be scattered across documentation or linked through outdated resources, forcing stakeholders to piece together the evidence needed for comparison \cite{biderman2024lessons, sokol2024benchmarkcards, zhao2025workflows}.

\para{Limited interpretability.}\ An excess of technical terminology can make evaluation reports difficult to understand, particularly for non-experts \cite{crisan2022interactive, liao2023designerly}. Varied reporting formats and insufficient contextual information can further hinder quick assessment of whether results apply to stakeholders' intended uses \cite{mitchell2019model, burnell2023rethink, sokol2024benchmarkcards}. Consequently, reported performance may offer insufficient guidance for choosing models under practical constraints \cite{ethayarajh-jurafsky-2020-utility, 10.1145/3706598.3713240}.

\smallskip
\noindent
$\rightarrow$ \textit{What we need:} All reports should be easy to understand for different stakeholders and come with clear guidelines on where such reports can be found.

\subsection{Accountability Issues}
\label{sec:accountability}  

Beyond informing research and model use, evaluation reporting supports algorithmic accountability by providing evidence to assess risks and question those responsible through suitable oversight mechanisms \cite{wieringa2020account, novelli2024accountability}.

\para{Limited risk assessment.}\ Incomplete reporting of safety evaluations and known limitations leaves stakeholders with insufficient evidence to scrutinize claims about model risks \cite{liu-etal-2023-responsible, reuel2025evaluates}, weakening assurances of model safety and the basis for oversight \cite{weidinger2024holistic, cen2024transparency}.

\para{Limited support for governance.}\ AI regulation and governance initiatives underscore the need for evaluation evidence that supports effective oversight \cite{laux2024three, reuel2024open}. Inconsistent reporting formats and indicators make this evidence difficult to compare across providers \cite{bommasani2024foundation}. Shared reporting requirements could help oversight bodies assess whether claims about models are adequately supported \cite{raji2020closing, mokander2024auditing}.

\para{Limited mechanisms for challenge and response.}\ Independent findings may fail to prompt investigation when channels for reporting concerns and procedures for handling them are unclear. Structured flaw reports and explicit response processes help connect external scrutiny to remediation \cite{cattell2024coordinated, mcgregor2025err, longpre2025position}.

\smallskip
\noindent
$\rightarrow$ \textit{What we need:} All model providers and deployers should report external risk assessments in alignment with governance requirements. Additionally, processes must be established to enforce such reporting formats and provide a clear route for questions or flaw reports. 

\section{Why Existing Artifacts Fall Short}
\label{sec:changeneed} 

The research community has long proposed documentation artifacts to improve transparency about machine learning artifacts or services. We examine four such artifacts, specifically focusing on their suitability for reporting GPAI model evaluation.

\para{Model Cards.}\ Model Cards \cite{mitchell2019model} are one of the earliest proposed artifacts with a focus on reporting all details about a model, including evaluation. A key gap with this framework is that its original guidance predates contemporary GPAI and centers on traditional ML evaluation. Although adaptable, it does not prescribe how each evaluated capability should be linked to its benchmark, prompting and metric configurations. Moreover, although the template includes ethical considerations, it does not provide a dedicated structure for reporting safety evaluations and risk assessments, which are central to GPAI evaluation \cite{weidinger2024holistic}.

\begin{figure*}[t]
\centering

\definecolor{ecRepro}{HTML}{315F91}
\definecolor{ecAccess}{HTML}{23776B}
\definecolor{ecAccount}{HTML}{79539A}

% Subheading followed by a short bullet
\newcommand{\ecEntry}[2]{%
    \textbf{#1}\\[2pt]
    {\footnotesize\textbullet~#2}%
}

\resizebox{\textwidth}{!}{%
\begin{tikzpicture}[
    item/.style={
        draw=black!20,
        rounded corners=2pt,
        text width=4.25cm,
        minimum height=11mm,
        inner xsep=2mm,
        inner ysep=2mm,
        align=left,
        font=\small
    },
    dimension/.style={
        item,
        minimum height=9mm,
        align=center,
        font=\small\bfseries,
        text=white
    },
    root/.style={
        item,
        text width=6cm,
        minimum height=10mm,
        align=center,
        fill=black!5,
        font=\large\bfseries
    },
    reproitem/.style={
        item,
        fill=ecRepro!6,
        draw=ecRepro!40
    },
    accessitem/.style={
        item,
        fill=ecAccess!6,
        draw=ecAccess!40
    },
    accountitem/.style={
        item,
        fill=ecAccount!6,
        draw=ecAccount!40
    },
    line/.style={
        draw=black!45,
        line width=0.65pt
    },
    arrow/.style={
        line,
        -{Latex[length=1.5mm]}
    }
]

% Framework
\node[root] (framework) at (0,1.7)
    {EvalCards Framework};

% Three dimensions
\node[dimension, fill=ecRepro, draw=ecRepro]
    (repro) at (-5.25,0) {Reproducibility};

\node[dimension, fill=ecAccess, draw=ecAccess]
    (access) at (0,0) {Accessibility};

\node[dimension, fill=ecAccount, draw=ecAccount]
    (account) at (5.25,0) {Accountability};

% Connect the framework to its dimensions
\coordinate (junction) at (0,0.95);
\draw[line] (framework.south) -- (junction);
\draw[arrow] (junction) -| (repro.north);
\draw[arrow] (junction) -- (access.north);
\draw[arrow] (junction) -| (account.north);

% Reproducibility
\node[reproitem] (r1) at (-5.25,-1.35) {
    \ecEntry{Modalities Evaluated}
            {Tested input/output modalities.}
};

\node[reproitem, below=2mm of r1] (r2) {
    \ecEntry{Languages Evaluated}
            {Languages actually evaluated on.}
};

\node[reproitem, below=2mm of r2] (r3) {
    \ecEntry{Capability Evaluations}
            {Benchmarks, metrics, and prompts.}
};

\node[reproitem, below=2mm of r3] (r4) {
    \ecEntry{Safety Evaluations}
            {Safety-specific evaluations.}
};

\node[reproitem, below=2mm of r4] (r5) {
    \ecEntry{Developer Footnotes}
            {Additional details for reproducing evaluations.}
};

% Accessibility
\node[accessitem] (a1) at (0,-1.35) {
    \ecEntry{Developer Accessibility}
            {Easy to write using templates.}
};

\node[accessitem, below=2mm of a1] (a2) {
    \ecEntry{Reader Accessibility}
            {Easy to read and interpret.}
};

\node[accessitem, below=2mm of a2] (a3) {
    \ecEntry{Discoverability}
            {Easy to find alongside models.}
};

% Accountability
\node[accountitem] (c1) at (5.25,-1.35) {
    \ecEntry{Risk Accountability}
            {Linked third-party risk assessments.}
};

\node[accountitem, below=2mm of c1] (c2) {
    \ecEntry{Governance Accountability}
            {Evolving support for governance needs.}
};

\node[accountitem, below=2mm of c2] (c3) {
    \ecEntry{Response Accountability}
            {Flaw reporting and responses.}
};

% Reproducibility branches
\coordinate (rbus) at ($(r5.west)+(-3mm,0)$);
\draw[line, draw=ecRepro!75]
    (repro.south) -- ++(0,-2mm) -| (rbus);

\foreach \n in {r1,r2,r3,r4,r5}{
    \draw[arrow, draw=ecRepro!75]
        (rbus |- \n.west) -- (\n.west);
}

% Accessibility branches
\coordinate (abus) at ($(a3.west)+(-3mm,0)$);
\draw[line, draw=ecAccess!75]
    (access.south) -- ++(0,-2mm) -| (abus);

\foreach \n in {a1,a2,a3}{
    \draw[arrow, draw=ecAccess!75]
        (abus |- \n.west) -- (\n.west);
}

% Accountability branches
\coordinate (cbus) at ($(c3.west)+(-3mm,0)$);
\draw[line, draw=ecAccount!75]
    (account.south) -- ++(0,-2mm) -| (cbus);

\foreach \n in {c1,c2,c3}{
    \draw[arrow, draw=ecAccount!75]
        (cbus |- \n.west) -- (\n.west);
}

\end{tikzpicture}%
}

\caption{The \evalcards{} framework and the key considerations
within its three dimensions.}
\label{fig:evalcards-framework}
\end{figure*}

\para{FactSheets.}\ FactSheets \cite{arnold2019factsheets} document AI services through supplier declarations. A key limitation for GPAI evaluation reporting is that their unit of documentation is an AI service, potentially combining multiple models. Fields concerning customer workflows, usage-data retention, and service monitoring depend on a particular deployment and may not apply when evaluating a GPAI model independently. Consequently, the framework requires adaptation to distinguish evaluation evidence about the underlying model from properties of the surrounding service and its operational context.

\para{Datasheets for Datasets.}\ Datasheets \cite{gebru2021datasheets} document how datasets are constructed and intended to be used. Their object of documentation is the dataset, rather than the GPAI model evaluated using it. They  do not provide a consolidated account of a model's evaluation conditions or performance across different evaluations.  

\para{BenchmarkCards.}\ BenchmarkCards \cite{sokol2024benchmarkcards} clarify what an LLM benchmark measures and how results should be interpreted . The organizing unit is the benchmark, whereas stakeholders also need an overview of a particular GPAI model's results across benchmarks. BenchmarkCards, however, can be valuable supporting artifacts to link
from GPAI evaluation reports, allowing users to investigate specific benchmark methods and limitations when needed.

Existing reporting artifacts thus do not fully address the evaluation reporting needs of GPAI models. The issues identified in our review motivate a dedicated framework that brings evaluation evidence together to support reproducibility, accessibility, and accountability. In the next section, we introduce EvalCards as a proposal for such a framework and discuss the key considerations that should guide its development and adoption.

\section{EvalCards: A Framework}
\label{sec:evalcards}

We identify issues in evaluation reporting and identify possible pathways that could address these issues as shown in \autoref{fig:evalfig}. Based on this, we introduce \evalcards{}, a framework for evaluation reporting for GPAI models as summarized in \autoref{fig:evalcards-framework}. Based on the issues identified in our critical synthesis, we organize the framework around three dimensions: \textit{reproducibility}, \textit{accessibility}, and \textit{accountability}. Within each dimension, we outline key considerations to guide the future design and development.

\subsection{The Reproducibility Dimension}
\label{sec:evalcards-reproducibility}

To ensure reproducibility of GPAI evaluations, we draw on the issues discussed earlier and identify the following information that reporting artifacts should contain to help researchers reconstruct evaluations and verify their results.

\para{Modalities evaluated.}\ \evalcards{} should state which input and output modalities were evaluated. For a model that accepts both images and text, for example, readers need
to know whether a reported result involved both modalities or text alone. This makes the input conditions needed to repeat the evaluation clear.

\para{Languages evaluated.}\ Any claims of multilingual support do not establish which languages were actually tested for the model. Reports should list the evaluated languages, allowing researchers to reproduce evaluations with the same language coverage.

\begin{figure*}[!tp]
\centering
\footnotesize
\setlength{\tabcolsep}{4pt}
\renewcommand{\arraystretch}{0.95}
\captionsetup{font=footnotesize,skip=3pt}
% Limit the card to 80% text width and 43% text height.
% The remaining half-page space accommodates the caption and float gap.
\newsavebox{\olmoEvalCardBox}
\sbox{\olmoEvalCardBox}{%
\resizebox{0.8\textwidth}{!}{%
\begin{evalcardv2}[OLMo-2-1124-7B-Instruct]

\vspace{0.2em}
\textbf{Modalities Evaluated}\vspace{0.2ex}

\rowcolors{2}{gray!10}{white}
\begin{tabularx}{\linewidth}{@{}>{\raggedright\arraybackslash}X>{\raggedright\arraybackslash}X@{}}
\toprule
\textbf{Type} & \textbf{Modalities} \\
\midrule
Input & Text \\
Output & Text \\
\bottomrule
\end{tabularx}

\vspace{0.3em}
\textbf{Languages Evaluated}\vspace{0.2ex}

\rowcolors{2}{gray!10}{white}
\begin{tabularx}{\linewidth}{@{}>{\raggedright\arraybackslash}X>{\raggedright\arraybackslash}X@{}}
\toprule
\textbf{Category} & \textbf{Languages} \\
\midrule
Total Number & 1 \\
List of Languages & English (\texttt{en}) \\
\bottomrule
\end{tabularx}

\vspace{0.3em}
\textbf{Capability Evaluations}\vspace{0.2ex}

\rowcolors{2}{gray!10}{white}
\begin{tabularx}{\linewidth}{@{}>{\raggedright\arraybackslash}p{3cm}>{\raggedright\arraybackslash}p{2cm}>{\raggedright\arraybackslash}p{2cm}>{\raggedright\arraybackslash}p{2cm}>{\raggedright\arraybackslash}X@{}}
\toprule
\textbf{Ability} &
\textbf{Benchmark} &
\textbf{Metric} &
\textbf{Zero-Shot} &
\textbf{Alternative Prompting} \\
\midrule
Knowledge &
\href{https://github.com/hendrycks/test}{MMLU} &
Exact Match & 61.3 (CoT) & N.A \\

Knowledge &
\href{https://github.com/AlexTMallen/adaptive-retrieval}{PopQA} &
Exact Match & N.A & 23.2 (15-Shot) \\

Knowledge &
\href{https://github.com/sylinrl/TruthfulQA}{TruthfulQA} &
Multiple Choice-2 & N.A & 56.5 (6-Shot) \\

Reasoning &
\href{https://github.com/suzgunmirac/BIG-Bench-Hard}{BBH} &
Exact Match & N.A & 51.4 (3-Shot CoT) \\

Reasoning &
\href{https://aclanthology.org/N19-1246/}{DROP} &
F1 & N.A & 60.5 (3-Shot) \\

Math &
\href{https://github.com/openai/grade-school-math}{GSM8K} &
Exact Match & N.A & 85.1 (8-Shot CoT) \\

Math &
\href{https://github.com/hendrycks/math}{MATH} &
Flex EM & N.A & 32.5 (4-Shot CoT) \\

Instruction Following &
\href{https://github.com/tatsu-lab/alpaca_eval}{AlpacaEval2} &
LC Winrate & 29.1 & N.A \\

Instruction Following &
\href{https://github.com/google-research/google-research/tree/master/instruction_following_eval}{IFEval}$^1$ &
Pass@1 & 72.3 & N.A \\
\bottomrule
\end{tabularx}

\vspace{0.3em}
\textbf{Safety Evaluations}\vspace{0.2ex}

\rowcolors{2}{gray!10}{white}
\begin{tabularx}{\linewidth}{@{}>{\raggedright\arraybackslash}p{3cm}>{\raggedright\arraybackslash}p{2cm}>{\raggedright\arraybackslash}p{2cm}>{\raggedright\arraybackslash}p{2cm}>{\raggedright\arraybackslash}X@{}}
\toprule
\textbf{Feature} &
\textbf{Benchmark} &
\textbf{Metric} &
\textbf{Zero-Shot} &
\textbf{Alternative Prompting} \\
\midrule
Safety &
\href{https://arxiv.org/abs/2411.15124}{T\"ulu 3 Safety}$^2$ &
N.A & 80.6 & N.A \\
\bottomrule
\end{tabularx}

\vspace{0.3em}
\textbf{Third-Party Risk Assessment}\vspace{0.2ex}

\rowcolors{2}{gray!10}{white}
\begin{tabularx}{\linewidth}{@{}>{\raggedright\arraybackslash}p{3cm}>{\raggedright\arraybackslash}X>{\raggedright\arraybackslash}p{3cm}>{\raggedright\arraybackslash}p{2cm}@{}}
\toprule
\textbf{Evaluator Identity} &
\textbf{Evaluated Risk} &
\textbf{Evaluation Result} &
\textbf{Report Link} \\
\midrule
N.A & N.A & N.A & N.A \\
\bottomrule
\end{tabularx}

\vspace{0.3em}
\textbf{Developer Footnotes}\vspace{0.2ex}

\rowcolors{2}{gray!10}{white}
\begin{tabularx}{\linewidth}{@{}>{\raggedright\arraybackslash}X@{}}
\toprule
$^1$ IFEval uses the ``loose'' evaluation criterion. \\
$^2$ T\"ulu 3 Safety combines multiple safety benchmarks;
see the linked paper for details. \\
\bottomrule
\end{tabularx}

\end{evalcardv2}%
}
}%
\ifdim\dimexpr\ht\olmoEvalCardBox+\dp\olmoEvalCardBox\relax>0.43\textheight
  \resizebox*{!}{0.43\textheight}{\usebox{\olmoEvalCardBox}}%
\else
  \usebox{\olmoEvalCardBox}%
\fi
\par
\caption{\evalcardd{} for \texttt{OLMo-2-1124-7B-Instruct}.}
\label{fig:evalcard-olmo}
\end{figure*}

\para{Capability evaluations.}\ For each capability evaluated, developers should identify the benchmark used and report the score together with its metric and prompting configuration. Zero-shot results and results from alternative strategies should be clearly distinguished, with the underlying prompts available. These details allow researchers to repeat the evaluation under the reported conditions and investigate differences in results.

\para{Safety evaluations.}\ Safety results should be similarly reported with a focus on safety-relevant issues.

\para{Developer footnotes.}\ This section provides space to document additional details of evaluation, including links to additional evaluation code or risk assessments from third parties. 

\subsection{The Accessibility Dimension}
\label{sec:evalcards-accessibility}

The accessibility dimension aims to ensure that evaluation reports remain accessible to all stakeholders. Any GPAI evaluation reporting artifact should commit to the following considerations and provide guidance for putting them into practice.

\para{Developer accessibility.}\ Reports should be \emph{easy to write}, and not mandate evaluations that can lead to undue compute requirements for model developers. Annotated templates and completed examples should guide authors through the required fields. Approaches such as \textit{CardGen} \cite{liu-etal-2024-automatic} could inform tools that draft report fields from technical documentation for author review. Guidelines for creating and updating reports should fit into existing model release workflows.

\para{Reader accessibility.}\ Reports should be \emph{easy to read} for stakeholders with different levels of technical expertise. For each named benchmark, reports could link to a completed \textit{ECBD Checklist} \cite{liu-etal-2024-ecbd} or \textit{BenchmarkCard} \cite{sokol2024benchmarkcards}, where available, giving readers further context if needed. Usability studies with intended readers should guide the presentation and language of future artifacts \cite{crisan2022interactive}.

\para{Discoverability.}\ Reports should be \emph{easy to find} wherever models are accessed. On Hugging Face, reports should be prominently linked from the model repository's model card. On GitHub, we recommend an \texttt{evalcard.md} file in the repository root, linked from the \texttt{README.md}. For models accessed through APIs, providers should link reports from developer documentation and dashboards.

\subsection{The Accountability Dimension}
\label{sec:evalcards-accountability}

The accountability dimension aims to ensure that any evaluation reporting artifact makes responsibility for its claims explicit.

\para{Risk accountability.}\ Beyond standard safety evaluations, reporting artifacts should accommodate third-party risk assessments \cite{longpre2025position}, naming the evaluator and linking to the external assessment for the tested model version.

\begin{figure*}[!tp]
\centering
\footnotesize
\setlength{\tabcolsep}{4pt}
\renewcommand{\arraystretch}{0.95}
\captionsetup{font=footnotesize,skip=3pt}
% Limit the card to 80% text width and 43% text height.
% The remaining half-page space accommodates the caption and float gap.
\newsavebox{\qwenEvalCardBox}
\sbox{\qwenEvalCardBox}{%
\resizebox{0.8\textwidth}{!}{%
\begin{evalcardv2}[Qwen3-4B-Base]

\vspace{0.2em}
\textbf{Modalities Evaluated}\vspace{0.2ex}

\rowcolors{2}{gray!10}{white}
\begin{tabularx}{\linewidth}{@{}>{\raggedright\arraybackslash}X>{\raggedright\arraybackslash}X@{}}
\toprule
\textbf{Type} & \textbf{Modalities} \\
\midrule
Input & Text \\
Output & Text \\
\bottomrule
\end{tabularx}

\vspace{0.3em}
\textbf{Languages Evaluated}\vspace{0.2ex}

\rowcolors{2}{gray!10}{white}
\begin{tabularx}{\linewidth}{@{}>{\raggedright\arraybackslash}X>{\raggedright\arraybackslash}X@{}}
\toprule
\textbf{Category} & \textbf{Languages} \\
\midrule
Total Number & 63 languages (65 language/script tags) \\
List of Languages &
af, ar, az, be, bg, bn, ca, cs, cy, da, de, el, en, es,
et, eu, fa, fi, fr, gu, he, hi, hr, hu, hy, id, it, ja,
ka, kk, kn, ko, lt, lv, mk, ml, mr, ms, ne, nl, no,
pa, pl, pt, ro, ru, sk, sl, so, sq, sr, sv, sw, ta,
te, th, tl, tr, uk, ur, uz, vi, zh, zh-Hans, zh-Hant \\
\bottomrule
\end{tabularx}

\vspace{0.3em}
\textbf{Capability Evaluations}\vspace{0.2ex}

\rowcolors{2}{gray!10}{white}
\begin{tabularx}{\linewidth}{@{}>{\raggedright\arraybackslash}p{3cm}>{\raggedright\arraybackslash}p{2cm}>{\raggedright\arraybackslash}p{2cm}>{\raggedright\arraybackslash}p{2cm}>{\raggedright\arraybackslash}X@{}}
\toprule
\textbf{Ability} &
\textbf{Benchmark} &
\textbf{Metric} &
\textbf{Zero-Shot} &
\textbf{Alternative Prompting} \\
\midrule
General Knowledge &
\href{https://huggingface.co/datasets/edinburgh-dawg/mmlu-redux}{MMLU-Redux} &
N.A & N.A & 77.3 \\

General Knowledge &
\href{https://github.com/idavidrein/gpqa}{GPQA-Diamond} &
Accuracy & N.A & 41.7 \\

Math &
\href{https://huggingface.co/datasets/HuggingFaceH4/MATH-500}{MATH-500} &
N.A & N.A & 84.8 \\

Math &
\href{https://artofproblemsolving.com/wiki/index.php/AIME_Problems_and_Solutions}{AIME-24} &
Accuracy & N.A & 25.0 \\

Math &
\href{https://artofproblemsolving.com/wiki/index.php/AIME_Problems_and_Solutions}{AIME-25} &
Accuracy & N.A & 19.1 \\

Reasoning &
\href{https://arxiv.org/abs/2502.01100}{ZebraLogic} &
N.A & N.A & 35.2 \\

Reasoning &
\href{https://github.com/8188zq/AutoLogi}{AutoLogi} &
N.A & N.A & 76.3 \\

Coding &
\href{https://livecodebench.github.io/}{LiveCode\allowbreak Bench v5} &
N.A & N.A & 21.3 \\

Coding &
\href{https://codeforces.com/}{CodeForces} &
Rating / Percentile & N.A & 842 / 33.7\% \\

Instruction Following &
\href{https://github.com/google-research/google-research/tree/master/instruction_following_eval}{IFEval} &
Strict prompt accuracy & N.A & 81.2 \\

Tool Integration &
\href{https://gorilla.cs.berkeley.edu/blogs/13_bfcl_v3_multi_turn.html}{BFCL v3} &
N.A & N.A & 57.6 \\
\bottomrule
\end{tabularx}

\vspace{0.3em}
\textbf{Safety Evaluations}\vspace{0.2ex}

\rowcolors{2}{gray!10}{white}
\begin{tabularx}{\linewidth}{@{}>{\raggedright\arraybackslash}p{3cm}>{\raggedright\arraybackslash}p{2cm}>{\raggedright\arraybackslash}p{2cm}>{\raggedright\arraybackslash}p{2cm}>{\raggedright\arraybackslash}X@{}}
\toprule
\textbf{Feature} &
\textbf{Benchmark} &
\textbf{Metric} &
\textbf{Zero-Shot} &
\textbf{Alternative Prompting} \\
\midrule
N.A & N.A & N.A & N.A & N.A \\
\bottomrule
\end{tabularx}

\vspace{0.3em}
\textbf{Third-Party Risk Assessment}\vspace{0.2ex}

\rowcolors{2}{gray!10}{white}
\begin{tabularx}{\linewidth}{@{}>{\raggedright\arraybackslash}p{3cm}>{\raggedright\arraybackslash}X>{\raggedright\arraybackslash}p{3cm}>{\raggedright\arraybackslash}p{2cm}@{}}
\toprule
\textbf{Evaluator Identity} &
\textbf{Evaluated Risk} &
\textbf{Evaluation Result} &
\textbf{Report Link} \\
\midrule
N.A & N.A & N.A & N.A \\
\bottomrule
\end{tabularx}

\vspace{0.3em}
\textbf{Developer Footnotes}\vspace{0.2ex}

\rowcolors{2}{gray!10}{white}
\begin{tabularx}{\linewidth}{@{}>{\raggedright\arraybackslash}X@{}}
\toprule
N.A \\
\bottomrule
\end{tabularx}

\end{evalcardv2}%
}
}%
\ifdim\dimexpr\ht\qwenEvalCardBox+\dp\qwenEvalCardBox\relax>0.43\textheight
  \resizebox*{!}{0.43\textheight}{\usebox{\qwenEvalCardBox}}%
\else
  \usebox{\qwenEvalCardBox}%
\fi
\par
\caption{\evalcardd{} for \texttt{Qwen3-4B-Base}.}
\label{fig:evalcard-qwen}
\end{figure*}

\para{Governance accountability.}\ Reports should accommodate governance-relevant evaluations as requirements and standards develop. For example, reports for GPAI models with systemic risk should include evidence from the model evaluations and adversarial testing required under Article~55 of the \textit{EU AI Act} \cite{act2024eu}. The framework should allow these reporting fields to evolve alongside governance guidance \cite{reuel2024open,reuel2024position}.

\para{Response accountability.}\ Reports should be accompanied by a clear procedure for challenging claims, linking to a submission form and identifying the team responsible for responding using an established schema such as the \textit{AI Flaw Report} \cite{longpre2025position}. Drawing on \textit{Coordinated Flaw Disclosure} \cite{cattell2024coordinated}, developers should publish review criteria and timelines for responses and decisions. Confirmed findings should lead to corrections or further evaluation.

\section{GPAI Case Studies}

In this section, we provide details about case studies we conducted with three GPAI models. The case studies serve two purposes: i) to examine existing reporting practices by identifying which information can be recovered and which remains missing or unclear form model evaluation reports; and ii) to provide concrete model-specific preliminary artifacts illustrating the minimum information that any \evalcardd{} artifact should report.

To examine how GPAI evaluations are reported in practice, we conduct three case studies of \texttt{OLMo-2-1124-7B-Instruct}, \texttt{Qwen3-4B-Base}, and \texttt{Gemini Flash 2.0}. For each model, we review available evaluation documentation across sources, including technical reports, model cards, release blogs, developer documentation, and linked evaluation repositories where available. We then attempt to compile the information into a preliminary \evalcardd{} format derived from the disclosure requirements identified in the \evalcards{} framework.  Information marked as unavailable (N.A) indicates that the relevant information for the evaluation was not available publicly.

\subsection{OLMo-2-1124-7B-Instruct}
\label{sec:case-olmo}

\autoref{fig:evalcard-olmo} presents the preliminary \evalcardd{} for \texttt{OLMo-2-1124-7B-Instruct}. We draw on the technical report \textit{2 OLMo 2 Furious}~\cite{olmo20252olmo2furious}, the model's \href{https://huggingface.co/allenai/OLMo-2-1124-7B-Instruct}{Hugging Face Model Card}, and the \href{https://github.com/allenai/olmes}{OLMES evaluation repository}.

\para{Reproducibility.}\ Evaluation metrics and prompt details were difficult to collect. We frequently had to cross-reference cited research papers and benchmark websites to establish which metric was used. Prompting conditions also varied substantially across benchmarks, limiting comparability across the reported evaluations.

\para{Accessibility.}\ Evaluation results were scattered across multiple tables in the main paper and appendix, making it difficult to obtain a concise overview of model capabilities. Although open-source models such as OLMo~2 generally provide more information than their closed counterparts, consolidating this information across remained time-consuming, limiting discoverability.

\begin{figure*}[!t]
\centering
\footnotesize
\setlength{\tabcolsep}{4pt}
\renewcommand{\arraystretch}{0.95}
\captionsetup{font=footnotesize,skip=3pt}
% Limit the card to 80% text width and 43% text height.
% The remaining half-page space accommodates the caption and float gap.
\newsavebox{\geminiEvalCardBox}
\sbox{\geminiEvalCardBox}{%
\resizebox{0.8\textwidth}{!}{%
\begin{evalcardv2}[Gemini Flash 2.0]

\vspace{0.2em}
\textbf{Modalities Evaluated}\vspace{0.2ex}

\rowcolors{2}{gray!10}{white}
\begin{tabularx}{\linewidth}{@{}>{\raggedright\arraybackslash}X>{\raggedright\arraybackslash}X@{}}
\toprule
\textbf{Type} & \textbf{Modalities} \\
\midrule
Input & Text, Image, Video, Audio \\
Output & Text \\
\bottomrule
\end{tabularx}

\vspace{0.3em}
\textbf{Languages Evaluated}\vspace{0.2ex}

\rowcolors{2}{gray!10}{white}
\begin{tabularx}{\linewidth}{@{}>{\raggedright\arraybackslash}X>{\raggedright\arraybackslash}X@{}}
\toprule
\textbf{Category} & \textbf{Languages} \\
\midrule
Total Number & N.A \\
List of Languages & N.A \\
\bottomrule
\end{tabularx}

\vspace{0.3em}
\textbf{Capability Evaluations}\vspace{0.2ex}

\rowcolors{2}{gray!10}{white}
\begin{tabularx}{\linewidth}{@{}>{\raggedright\arraybackslash}p{3cm}>{\raggedright\arraybackslash}p{2cm}>{\raggedright\arraybackslash}p{2cm}>{\raggedright\arraybackslash}p{2cm}>{\raggedright\arraybackslash}X@{}}
\toprule
\textbf{Ability} &
\textbf{Benchmark} &
\textbf{Metric} &
\textbf{Zero-Shot} &
\textbf{Alternative Prompting} \\
\midrule
General Knowledge &
\href{https://github.com/TIGER-AI-Lab/MMLU-Pro}{MMLU-Pro} &
N.A & N.A & 76.6 \\

Factual Knowledge &
\href{https://www.kaggle.com/benchmarks/google/facts-grounding}{FACTS Grounding} &
N.A & N.A & 83.6 \\

Code &
\href{https://arxiv.org/html/2403.05530v5}{Natural2\allowbreak Code} &
N.A & N.A & 92.9 \\

Code &
\href{https://bird-bench.github.io/}{BIRD-SQL Dev} &
N.A & N.A & 56.9 \\

Code &
\href{https://livecodebench.github.io/}{LiveCode\allowbreak Bench} &
N.A & N.A & 35.1 \\

Math &
\href{https://github.com/hendrycks/math}{MATH} &
N.A & N.A & 89.7 \\

Math &
\href{https://arxiv.org/html/2403.05530v5}{HiddenMath} &
N.A & N.A & 63 \\

Reasoning &
\href{https://github.com/idavidrein/gpqa}{GPQA-Diamond} &
N.A & N.A & 62.1 \\

Long Context &
\href{https://arxiv.org/html/2403.05530v5}{MRCR (1M)} &
N.A & N.A & 69.2 \\

Image &
\href{https://mmmu-benchmark.github.io/}{MMMU} &
N.A & N.A & 70.7 \\

Image &
\href{https://github.com/reka-ai/reka-vibe-eval}{Vibe-Eval (Reka)} &
N.A & N.A & 56.3 \\

Video &
\href{https://egoschema.github.io/}{EgoSchema (Test)} &
N.A & N.A & 71.5 \\

Audio &
\href{https://github.com/facebookresearch/covost}{CoVoST2 (21 Lang)} &
N.A & N.A & 39.2 \\
\bottomrule
\end{tabularx}

\vspace{0.3em}
\textbf{Safety Evaluations}\vspace{0.2ex}

\rowcolors{2}{gray!10}{white}
\begin{tabularx}{\linewidth}{@{}>{\raggedright\arraybackslash}p{3cm}>{\raggedright\arraybackslash}p{2cm}>{\raggedright\arraybackslash}p{2cm}>{\raggedright\arraybackslash}p{2cm}>{\raggedright\arraybackslash}X@{}}
\toprule
\textbf{Feature} &
\textbf{Benchmark} &
\textbf{Metric} &
\textbf{Zero-Shot} &
\textbf{Alternative Prompting} \\
\midrule
N.A & N.A & N.A & N.A & N.A \\
\bottomrule
\end{tabularx}

\vspace{0.3em}
\textbf{Third-Party Risk Assessment}\vspace{0.2ex}

\rowcolors{2}{gray!10}{white}
\begin{tabularx}{\linewidth}{@{}>{\raggedright\arraybackslash}p{3cm}>{\raggedright\arraybackslash}X>{\raggedright\arraybackslash}p{3cm}>{\raggedright\arraybackslash}p{2cm}@{}}
\toprule
\textbf{Evaluator Identity} &
\textbf{Evaluated Risk} &
\textbf{Evaluation Result} &
\textbf{Report Link} \\
\midrule
N.A & N.A & N.A & N.A \\
\bottomrule
\end{tabularx}

\vspace{0.3em}
\textbf{Developer Footnotes}\vspace{0.2ex}

\rowcolors{2}{gray!10}{white}
\begin{tabularx}{\linewidth}{@{}>{\raggedright\arraybackslash}X@{}}
\toprule
N.A \\
\bottomrule
\end{tabularx}

\end{evalcardv2}%
}
}%
\ifdim\dimexpr\ht\geminiEvalCardBox+\dp\geminiEvalCardBox\relax>0.43\textheight
  \resizebox*{!}{0.43\textheight}{\usebox{\geminiEvalCardBox}}%
\else
  \usebox{\geminiEvalCardBox}%
\fi
\par
\caption{\evalcardd{} for \texttt{Gemini Flash 2.0}.}
\label{fig:evalcard-gemini}
\end{figure*}

\para{Accountability.}\ Safety performance was reported as a single average across six datasets, without disaggregated scores. This prevented us from determining how the model performed on specific risks, such as toxicity, bias, or hallucination. The absence of per-dataset results limited meaningful assessment and comparison of the model's safety profile. No risk assessment was reported and no process for flaw reporting was available anywhere for chalenging evaluations.

\subsection{Qwen3-4B-Base}

\autoref{fig:evalcard-qwen} presents an \evalcardd{} for \texttt{Qwen3-4B-Base}. Released on 29 April 2025 under the Apache 2.0 license, the model is available through platforms including \href{https://huggingface.co/Qwen/Qwen3-4B}{Hugging Face}, \href{https://modelscope.cn/organization/qwen}{ModelScope}, and \href{https://www.kaggle.com/models/qwen-lm/qwen-3}{Kaggle}. 

\para{Reproducibility.}\
The \href{https://qwenlm.github.io/blog/qwen3/}{release blog} reported training coverage of 119 languages without clearly identifying which languages had been evaluated. Establishing evaluation coverage required examining the languages represented in the cited benchmarks separately. The blog and Hugging Face repository described capabilities in creative writing, role-playing, multi-turn dialogue, instruction following, and reasoning without clearly connecting each claim to the benchmarks or datasets used to evaluate it. Metrics and prompting strategies were also unavailable for many benchmark scores reported in the blog. Because we could not establish whether these results used zero-shot prompting, we placed them in the alternative prompting column.

\para{Accessibility.}\
Only a limited set of evaluations was available in the release blog during the first week after release, restricting the information available for informed use. This inhibits discoverability at the point of model release.

\para{Accountability.}\
We did not find specific safety evaluations in the documentation examined, despite the model having been released. Our card also lacks an identifiable third-party evaluator and a linked risk assessment for this model version, leaving risk accountability unestablished in the collected material. The \href{https://github.com/QwenLM/Qwen3}{Qwen repository} provides community contact routes, but its README does not specify a structured flaw-reporting process.

This preliminary artifact represents one instantiation of the EvalCards framework. Putting reporting standards into practice requires accompanying measures that make reports readily available where models are accessed and interpretable to relevant stakeholders. To support accountability, artifacts should also link to clear flaw-reporting mechanisms, identify those responsible for handling concerns, and specify response timelines and procedures for communicating corrections. The artifact should therefore be developed alongside the practices needed to keep evaluation claims accessible, open to scrutiny, and responsive to identified shortcomings.

\subsection{Gemini Flash 2.0}

\autoref{fig:evalcard-gemini} presents an \evalcardd{} for \texttt{Gemini Flash 2.0}. The model was accessible through the Gemini API and described in the official Developer Studio documentation\footnote{\url{https://ai.google.dev/gemini-api/docs/models}} and a December 2024 blog post.\footnote{\url{https://blog.google/technology/google-deepmind/google-gemini-ai-update-december-2024/\#gemini-2-0}} 

\para{Reproducibility.} \ The \href{https://ai.google.dev/gemini-api/docs/models#supported-languages}{Developer Studio documentation} listed languages the model was ``trained to work on'' without specifying whether performance in those languages had been evaluated. We could not establish the evaluation metric or prompting strategy for the benchmark scores, thus placing these scores in the alternative prompting column.

\para{Accessibility.} \ Evaluation information was distributed between the Developer Studio documentation and the release blog, requiring readers to consult multiple sources to assemble an overview of the model's evaluated capabilities. Missing explanations of metrics and prompting conditions also limited reader accessibility by making scores difficult to interpret without investigating the benchmarks.

\para{Accountability.}
The release blog described a review by Google's Responsibility and Safety Committee (RSC) to identify and understand potential risks. However, that blog did not disclose concrete safety findings, detailed red-teaming procedures, or the benchmarks used for that review. Its reference to a safety review therefore did not provide the evaluation evidence needed to assess the model's safety performance. For response accountability, the \href{https://ai.google.dev/gemini-api/docs/troubleshooting}{API troubleshooting guidance} points users to a developer forum. Neither that guidance nor the model card specifies a structured flaw-reporting procedure for disputing published evaluation claims.

\section{Discussion}
\label{sec:futurework}

The EvalCards framework translates common evaluation reporting issues into a set of considerations for future reporting artifacts, connecting what should be reported with how stakeholders can find, understand, and question that information. Its practical value will depend on whether these considerations can be implemented in formats that support stakeholders while keeping reporting effort manageable. The preliminary format presented here provides a starting point for this work, but its usefulness remains to be established through empirical validation. We therefore discuss priorities for refining artifact design and evaluating its use in practice.

\para{Artifact Design.} We provide a preliminary artifact format here based on the fields identified in the reproducibility dimension (\autoref{sec:evalcards-reproducibility}). This is not intended as a final version and further work should consider additional fields for reporting such as latency \cite{husom2025sustainable}, compute cost \cite{wang2024reasoning, mcfadyen2026inference}, or environmental impact \cite{hershcovich-etal-2022-towards}. To support standardization, artifacts can enforce guidelines on the use of benchmark identifiers and versions \cite{biderman2024lessons} and consistent language labels using standards such as \textit{ISO 639} \cite{dalby-etal-2004-standards,gillis-webber-tittel-2020-framework}. 

\para{Stakeholder Validation.} Participatory design and user studies should refine the format with developers and downstream users \cite{crisan2022interactive,liao2023designerly}. Comparing concise and detailed versions could establish whether stakeholders can locate evaluation conditions, recognize unsupported claims, and identify how to challenge them. Studies should measure reporting effort alongside reader comprehension. Longer-term pilots should test whether reports remain usable across updates, informed by maintenance problems documented in model leaderboards \cite{zhao2025workflows}.

Our main argument is one for a shift in norms: evaluation reporting is not a marketing exercise but a core component of what it means to release a model responsibly.  While the broader challenge of how to evaluate models remains open and complex \cite{laskar2024systematic, chang2024survey, gao2025llm}, our focus here is narrower but nonetheless critical: improving how evaluations are reported. We hope this work helps move the field toward a culture of improved evaluation reporting standards for GPAI models.

\section{Conflict of interest}
Authors state no conflict of interest.

% Include the complete bibliography, as in the NMI draft.
\clearpage
\nocite{*}
\bibliography{custom}
\end{document}